\documentclass[twoside,11pt]{article}

\usepackage{jmlr2e}
\usepackage{times,epsfig,graphics,amssymb,latexsym,amsmath,algorithm,algorithmic,subcaption,setspace}
\usepackage{array,url}
\usepackage[usenames]{color}
\usepackage{enumitem}
\usepackage{multirow}
\usepackage{natbib}

\newcommand\indep{\protect\mathpalette{\protect\independenT}{\perp}}
\def\independenT#1#2{\mathrel{\rlap{$#1#2$}\mkern2mu{#1#2}}}

\def\diag{\mathop{\rm diag}}

\newtheorem{Assumption}{Assumption}
\newtheorem{cor}{Corollary}
\newtheorem{lem}{Lemma}

\newcommand {\bfepsilon} {\mbox{\boldmath $\epsilon$}}

\def\bx{\mathop{\bf x}}
\def\bX{\mathop{\bf X}}

\def\bB{\mathop{\bf B}}
\def\bA{\mathop{\bf A}}

\def\be{\mathop{\bf e}}
\def\bI{\mathop{\bf I}}

\def\bM{\mathop{\bf M}}

\ShortHeadings{Learning linear non-Gaussian DAG}{Zhao and He and Wang}
\firstpageno{1}

\begin{document}

\title{Learning linear non-Gaussian directed acyclic graph with diverging number of nodes}

\author{\name Ruixuan Zhao \email  ruixuzhao2-c@my.cityu.edu.hk\\
	\addr School of Data Science\\
	City University of Hong Kong\\
	Kowloon Tong, Kowloon, Hong Kong
	\AND
	\name Xin He \email he.xin17@mail.shufe.edu.cn \\
	\addr School of Statistics and Management\\
	Shanghai University of Finance and Economics \\
	Shanghai, China
	\AND
	\name Junhui Wang \email j.h.wang@cityu.edu.hk \\
	\addr School of Data Science\\
	City University of Hong Kong\\
	Kowloon Tong, Kowloon, Hong Kong
}

\editor{ }

\maketitle

\begin{abstract}
Acyclic model, often depicted as a directed acyclic graph (DAG), has been widely employed to represent directional causal relations among collected nodes. In this article, we propose an efficient method to learn  linear non-Gaussian DAG in high dimensional cases, where  the noises can be of any continuous non-Gaussian  distribution. This is in sharp contrast to most existing DAG learning methods assuming Gaussian noise with additional variance assumptions to attain exact DAG recovery. The proposed method leverages a novel concept of topological layer to facilitate the DAG learning. Particularly, we show that the topological layers  can be exactly reconstructed  in a bottom-up fashion,  and   the parent-child relations among nodes in each layer can  also be consistently established. More importantly, the proposed method does not require  the faithfulness or parental faithfulness assumption which has been widely assumed in the literature of DAG learning. Its advantage is also supported by the numerical comparison against some popular competitors in various simulated examples as well as a real application on the  global spread of COVID-19.
\end{abstract}


\section{Introduction}	

Directed acyclic graph (DAG) provides an elegant way to represent directional or causal structures among collected nodes, which finds applications in a broad variety of domains, including  genetics \citep{Sachs2005}, finance \citep{Sanford2012} and social science \citep{Newey1999}. In recent years, learning the DAG structures from observed data has attracted tremendous attention from both academia and industries.  

In literature, various structure learning methods have been proposed to recover the Markov equivalence class \citep{Spirtes2000, Peters2017} of a DAG, which can be roughly categorized into three classes. The first class is the constraint-based method \citep{Spirtes2000, Kalisch2007}, which uses some  local conditional independence criterion to test pairwise causal relations. The second class, referred as the score-based method \citep{Chickering2002, Zheng2018}, attempts to optimize some goodness-of-fit measures among the possible  graph space. The last class  \citep{Tsamardinos2006, Nandy2018} combines the constraint-based method and score-based method. Although success has been widely reported, most aforementioned methods  can only recover the Markov equivalence class and their computational burden remains a severe bottleneck. Recently, efforts have been made to pursuit exact DAG recovery. Specifically, \cite{Peters2014} shows that  a linear Gaussian DAG is identifiable under the equal variance assumption, and \cite{Ghoshal2018} and \cite{ Park2020} relax the Gaussianity assumption but still require an explicit order among noise variances. Under these assumptions, a number of learning methods are proposed to recovery the exact DAG structure  \citep{ChenW2019, Yuan2019, Shen2020, Park2021}, yet these assumptions of Gaussianity or ordered noise variances are often difficult to verify in practice.

Linear non-Gaussian DAG, also known as linear non-Gaussian acyclic model (LiNGAM) in literature \citep{Shimizu2006}, relaxes the Gaussianity assumption, and its identifiability does not require any additional noise variance assumption. It is clear that linear non-Gaussian DAG can accommodate more flexible distributions, yet it has only received limited attention in literature \citep{Shimizu2006,Shimizu2011, Hyvarinen2013,Wang2020}. Specifically, \cite{Shimizu2006}  proposes an iterative search  algorithm to recover the  causal ordering  of  linear non-Gaussian DAG by using linear independent component analysis (ICA) and permutation.  Subsequently, \cite{Shimizu2011} proposes a multiple-step algorithm to  learn the linear non-Gaussian DAG by some pairwise statistics, which is further extended in \cite{Hyvarinen2013} to iteratively identify pairwise causal ordering by  likelihood ratio tests. The statistical properties of these methods remain  largely unknown, not to mention their expensive computational cost.	Most recently, \cite{Wang2020} proposes a modified direct learning algorithm for linear non-Gaussian DAG in high dimensional cases, which   sequentially recovers the causal ordering  with a moment-based criterion and reconstructs the directed structure with hard-thresholding. Yet,  it requires the parental faithfulness condition and its computational complexity is of exponential order of the maximum in-degree.

In this paper, we propose a novel method to learn linear non-Gaussian DAG in high dimensional cases based on a novel concept of topological layer. It assures that any DAG can be reformulated into a unique topological structure with  $T$ layers, where the parents of a node must belong to its upper layers, and thus acyclicity is naturally guaranteed.  More importantly,  we show that the topological layers  can be exactly reconstructed via precision matrix estimation and independence testing procedure in a bottom-up fashion,  and   the parent-child relations can  be  directly obtained  from the estimated precision matrix.  {These results are obtained without requiring the popular faithfulness \citep{Uhler2013,Peters2017} or parental faithfulness assumption \citep{Wang2020}}.  The constructive proof also motivates an  efficient learning algorithm for the proposed method,  whose complexity is  much smaller than most existing linear non-Gaussian DAG learning methods \citep{Shimizu2006,Shimizu2011, Wang2020}.

The main contribution of this paper is the development of a  novel and efficient method to learn linear non-Gaussian DAG in high dimensional cases, and the investigation on its statistical guarantee in terms of exact causal structure recovery.   	More precisely,  we show that the topological layers of the DAG can be exactly reconstructed in  Theorem \ref{thm:A0} and Corollary \ref{cor:layers}, and the parent-child relations can  be  directly recovered along with the topological layers in Corollary \ref{cor:parents}. 	We  connect learning method and  precision matrix estimation by proving that   the topological layers  can be exactly reconstructed via precision matrix estimations in a bottom-up fashion,  and the parent-child relations  can be obtained directly from the obtained precision matrix. The statistical guarantees of  the proposed method by using  graphical Lasso  and distance covariance measure is established  with sub-Gaussian and ($4m$)-th bounded moment noise distributions, respectively. The established  consistency results are governed by the sample size, the number of nodes, and the maximum cardinality of Markov blankets \citep{Peters2017}. Most interestingly, the obtained results allow the number of nodes and the maximum cardinality of Markov blankets to diverge with the sample size at some fast rate, which is particularly attractive in high-dimensional learning method.

The rest of this paper is organized as follows.  Section \ref{sec:prel} introduces some background of  linear non-Gaussian DAG. Section \ref{sec:iden} introduces the  concept of topological layers and shows that the topological layers  and the parent-child relations can be exactly reconstructed in  a bottom-up fashion. Section \ref{sec:alg} provides an efficient learning algorithm for linear non-Gaussian DAG in high dimensional cases, and Section \ref{sec:theory} establishes the reconstruction consistency of the proposed method  under mild conditions.  Numerical experiments  on several simulated examples and one real application to the spread of COVID-19 are conducted in Section \ref{sec:num}.  Section \ref{sec:sum} contains a brief discussion, and all the technical details are provided in Appendix.

\section{Preambles}\label{sec:prel}

Consider a DAG ${\cal G}=\{ {\cal N},{\cal E}\}$, encoding the joint distribution $P(\bx)$ of $\bx=(x_1,...,x_p)^T \in {\cal R}^p$, where  ${\cal N}=\{1,\ldots,p\}$ consists of a set of nodes associated with each coordinate of $\bx$, and  $ {\cal E}\subset {\cal N}\times{\cal N}$ consists of all the directed edges among the nodes. The directed edge from node $j$ to node $k$ is denoted as $j \rightarrow k$, indicating their parent-child relationship.  For simplicity, we denote node $k$'s parents as $\mbox{pa}_k$, its children as $\text{ch}_k$, its descendants as $\text{de}_k$, its non-descendants as $\text{nd}_k$, and its Markov blanket as $\mbox{mb}_k=\mbox{pa}_k \cup \mbox{ch}_k \cup \{i\in \mbox{pa}_j \backslash \{k\} | j \in \mbox{ch}_k\}$. It is also assumed that ${\cal G}$ satisfies the Markov property \citep{Spirtes2000}, and thus ${P}(\bx)$ can be factorized as ${P}(\bx)=\prod_{k=1}^pP(x_k|{\bx}_{\mbox{pa}_k})$,  where ${\bx}_{\mbox{pa}_k} = \{ x_j : j \in \mbox{pa}_k \}$. 

Once each node $x_k$ is centered with mean zero, the graph structure in ${\cal G}$ can be embedded into a linear structural equation model (SEM),
\begin{align}\label{eqn:1}
	x_k=\sum_{j\in \mbox{pa}_k} {\beta_{kj}}x_j + \epsilon_k; \ k=1,...,p,
\end{align}
where ${\beta_{kj}}\neq 0$ for any $j \in \mbox{pa}_k$,  $\epsilon_k$ denotes a continuous non-Gaussian noise with variance $\sigma_k^2$, and $\epsilon_l\indep \epsilon_k$ for any $l \neq k$. This independent noise condition further implies that $\epsilon_k \indep x_l$ for any $l \notin \mbox{de}_k\cup \{k\}$. It is also often assumed that there is no unobserved confounding effect among the observed nodes in $\cal G$, which is known as the casual sufficiency condition  in literature \citep{Spirtes2000}. 

Note that the SEM model in \eqref{eqn:1} can be organized into a matrix form
\begin{align*}
	\bx=\bB\bx+{\bfepsilon},
\end{align*}
where ${\bB}=(\beta_{kj})_{k,j}\in {\cal R}^{p\times p}$ and $\bfepsilon=(\epsilon_1,...,\epsilon_p)^T$ is the noise vector with covariance matrix $\mathbf{\Omega}=\diag \{ \sigma_1^2,...,\sigma_p^2 \}$. Simple algebra yields that
\begin{align}\label{inv1}
	\bx =(\bI-{\bB})^{-1}\bfepsilon ={\bA}\bfepsilon,
\end{align}
and $x_k=\sum_{j=1}^p a_{kj}\epsilon_j$, where $a_{kj}$ is  the $(k,j)$-th element of $\bA=(\bI-{\bB})^{-1}$, representing the total effect of  $x_j$ on $x_k$.  The SEM model implies that $\epsilon_j \indep x_k$ and thus  $a_{kj}=0$ for any node $j \in \mbox{de}_k$.  Moreover, the covariance matrix of $\bx$ is $\mathbf{\Sigma} = (\bI-\bB)^{-1} \mathbf{\Omega} (\bI-\bB)^{-T}$, and the corresponding  precision matrix is 
$\mathbf{\Theta}=\mathbf{\Sigma}^{-1}=(\bI-\bB)^{T} \mathbf{\Omega}^{-1}(\bI-\bB)$.  In the sequel,  we use ${\Theta}_{lk}$ to denote the $(l,k)$-th element of $\mathbf \Theta$, and ${\mathbf \Theta}_{-ll}$ to denote the $l$-th column of $\mathbf{\Theta}$ without $\Theta_{ll}$.


\section{Topological layers}\label{sec:iden}

In this section, we introduce a novel concept of topological layer, which allows us to convert a DAG into a unique topological structure. Particularly, given a DAG ${\cal G}$, we construct its topological structure by assigning each node to one and only one layer, based on its longest distance to one of the leaf nodes.  

\begin{figure}[!h]
	\centering
	\includegraphics[width=0.7\textwidth]{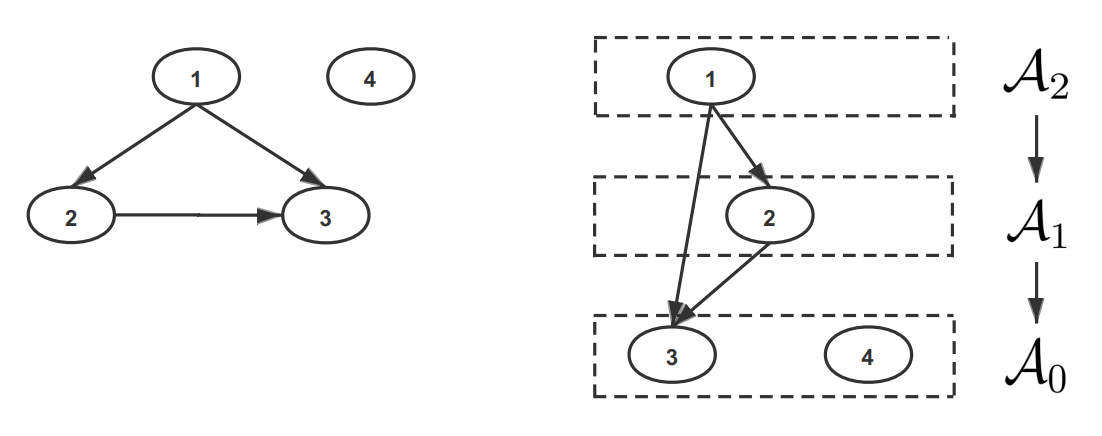}
	\caption{A toy DAG and its topological structure with three layers.}\label{fig:0}	
\end{figure}

Without loss of generality, we assume ${\cal G}$ has a total of $T$ layers,  and ${\cal A}_t$ denotes all the nodes contained in the $t$-th layer, for $t = 0, \ldots, T-1$.  It is clear that  $\cup_{t=0}^{T-1}{\cal A}_{t}={\cal N}$, and all the leaf nodes and isolated nodes in ${\cal G}$ belong to the lowest layer ${\cal A}_{0}$. For each node $k\in {\cal A}_t$, it follows from the layer construction that {$\mbox{pa}_k \subset {\cal S}_{t+1}=\cup_{d=t+1}^{T-1}{\cal A}_d$}, and thus  acyclicity  is automatically guaranteed. Note that ${\cal S}_0={\cal N}$.  Figure \ref{fig:0} illustrates a toy DAG in the left panel, and its converted topological structure with three layers in the right panel.

In  Figure \ref{fig:0}, node $4$ is an isolated node and node $3$ has no child node, and thus they both belong to ${\cal A}_{0}$. Node 3 has two parent nodes, where node 2 belongs to ${\cal A}_1$ but node 1 belongs to ${\cal A}_2$ due to the existence of a longer path $1 \rightarrow 2 \rightarrow 3$.  It is clear that the concept of topological layer is general and it can restructure any DAG in such a way that causal ordering among each layers is uniquely determined. This is in sharp contrast to the idea of causal ordering in literature \citep{Shimizu2006, Shimizu2011, Wang2020}, which only requires that each node is ranked behind its parents.  For  the toy DAG in Figure \ref{fig:0}, it induces multiple possible causal orderings among the four nodes, such as $1 \rightarrow 2 \rightarrow 3 \rightarrow 4$, $1 \rightarrow 2 \rightarrow 4 \rightarrow 3$, $1 \rightarrow 4 \rightarrow 2 \rightarrow 3$, or $4 \rightarrow 1 \rightarrow 2 \rightarrow 3$. This indeterministic causal ordering may cause unnecessary  estimation instability and computational inefficiency in  reconstructing the DAG  structures.

\subsection{Reconstruction of linear non-Gaussian DAG}\label{sec:recon}

To be self-contained, we first restate the Darmois-Skitovitch theorem \citep{Darmois1953, Skitovitch1953}, which is crucial for the reconstruction of  topological layers of a linear non-Gaussian DAG.

\begin{lem}\label{lem1}
	(Darmois-Skitovitch, 1953) Define two random variables $u_1$ and $u_2$ as linear combinations of independent random variables $s_i, i=1,...,m$, that $u_1=\sum_{i=1}^m c_{1,i} s_i ~\mbox{and}~ u_2=\sum_{i=1}^m c_{2,i} s_i.$
	Then, if $u_1$ and $u_2$ are independent, all variables $s_i$ with $c_{1,i}c_{2,i}\neq 0$  are Gaussian distributed.
\end{lem}

Lemma \ref{lem1} shows that if $s_i$'s are non-Gaussian distributed,  it is impossible to construct two independent linear combinations of $s_i$'s. This fact motivates us to use independence test to identify nodes in each layers in a bottom-up fashion.

\begin{theorem} \label{thm:A0}
	Suppose that $\bx=(x_1,..,x_p)^T \in {\cal R}^{p}$ is generated from the linear SEM model in \eqref{eqn:1} with precision matrix $\mathbf{\Theta}$. For any $l \in {\cal N}$,  we regress $x_l$ on all other nodes $\bx_{{\cal N}\backslash \{l\}}$, and denote the residual as $e_{l, {\cal N}}=x_l-\bx^T_{{\cal N}\setminus \{l\}}\bM_{\cal N}^{(l)}$, where $\bM_{\cal N}^{(l)}=-{\mathbf{\Theta}}_{-ll}/{{\Theta}}_{ll}$.  Then, we have $l \in {\cal A}_{0}$ if and only  if $ e_{l, {\cal N}} \indep x_k$ for any $k\in { {\cal N}\backslash \{l\}  }$.  
\end{theorem}

Theorem \ref{thm:A0}  provides a sufficient and necessary condition to identify nodes in ${\cal A}_{0}$.  After all the nodes in ${\cal A}_0$ are identified, we can remove them from $\cal N$ and denote ${\cal S}_1={\cal N}\setminus  {\cal A}_0$. Next, we apply similar treatment to ${\cal S}_1$ as in Theorem 1 to identify ${\cal A}_1$, and then  ${\cal A}_2$, until all nodes are assigned to layers. Denote $\mathbf{\Theta}^{{\cal S}_t}$ as the precision matrix of $\bx_{{\cal S}_t}$ and $\mathbf{\Theta}^{{\cal S}_{0}}= \mathbf{\Theta}$. Further, denote $[{\Theta}^{{\cal S}_t}]_{lk}$  as the element of $\mathbf{\Theta}^{{\cal S}_t}$ corresponding to nodes $l$ and $k$, and $[\mathbf{\Theta}^{{\cal S}_t}]_{-ll}$ as the column of $\mathbf{\Theta}$ corresponding to node $l$ without $[{\Theta}^{{\cal S}_t}]_{ll}$. Corollary \ref{cor:layers} summarizes the reconstruction of  all the topological layers for a linear non-Gaussian DAG.

\begin{cor}\label{cor:layers}
	Suppose that all the conditions in Theorem \ref{thm:A0} are satisfied and the layers ${\cal A}_0,...,{\cal A}_{t-1}$ have been reconstructed. For any $l \in {\cal S}_t$,  we regress $x_l$ on $\bx_{{\cal S}_t\backslash\{l\}}$ and denote the residual as $e_{l, {\cal S}_t}=x_l-\bx_{{\cal S}_t\backslash \{l\}}^T \bM_{{\cal S}_t}^{(l)}$ with $\bM_{{\cal S}_t}^{(l)}=-[\mathbf{\Theta}^{{\cal S}_t}]_{-ll}/[{\Theta}^{{\cal S}_t}]_{ll}$. Then we have $l \in {\cal A}_{t}$ if and only if $e_{l, {\cal S}_t}\indep x_k$ for any $k\in { {\cal S}_t\backslash \{l\}  }$.  
\end{cor}	

The proof of Corollary \ref{cor:layers} is similar to that of Theorem \ref{thm:A0} with slight modification by replacing ${\cal N}$ with ${\cal S}_t$. The reconstruction of the topological layers follows immediately by applying Corollary \ref{cor:layers} repeatedly in the sense of mathematical induction. Furthermore, the parent sets of each node in the DAG can be determined during the reconstruction of the topological layers as well.

\begin{cor} \label{cor:parents}
	Suppose that all the conditions in Theorem \ref{thm:A0} are satisfied. For  any $l \in {\cal A}_{t}$ and $k \in {\cal S}_{t+1}$, we have $\beta_{lk}=-[{\Theta}^{{\cal S}_{t}}]_{lk}/[{\Theta}^{{\cal S}_{t}}]_{ll}$, and thus $\mbox{pa}_l=\{ k \in {\cal S}_{t+1}: [{\Theta}^{{\cal S}_{t}}]_{lk} \neq 0\}$.
\end{cor}

Corollary \ref{cor:parents} follows immediately after Corollary \ref{cor:layers}, and assures that the parent-child relations can also be  sequentially reconstructed along with the topological layers. It is  important to point out that both Corollaries \ref{cor:layers} and \ref{cor:parents}   don't require  the popularly-adopted faithfulness assumption \citep{Uhler2013, Peters2017} or the parental faithfulness assumption \citep{Wang2020}.

\subsection{An illustrative example}\label{sec:toy}

Consider a simple linear non-Gaussian DAG generated as follows,
\begin{align}\label{sec2:toy}
	x_1=\epsilon_1, \ \ x_2=\beta_{21}x_1+\epsilon_2, \ \   x_3=   \beta_{31}x_1 + \beta_{32}x_2+\epsilon_3 , \ \mbox{and}~ x_4=\epsilon_4,
\end{align}
where  $\epsilon_l$ is a non-Gaussian distributed noise, and $\epsilon_l \indep \epsilon_k$ for any $l \neq k$. Clearly, both the faithfulness and the parental faithfulness assumptions are violated when $\beta_{32}\beta_{21}+\beta_{31}=0$. Yet such a linear non-Gaussian DAG can be successfully reconstructed with the proposed criteria in Section \ref{sec:recon}.

We first regress $x_l$ on ${\bx}_{-l}$ and obtain the residuals $e_{l, {\cal N}}$ for $l=1,\ldots,4$. It is easy to see that $e_{3, {\cal N}}=\epsilon_3$ and $e_{4, {\cal N}}=\epsilon_4$, and each of them  is  independent with all the other three nodes. For $e_{1, {\cal N}}$ and $e_{2, {\cal N}},$ it follows from similar treatment as in the proof of Theorem 1 that
\begin{align*}
	e_{1, {\cal N}} &=\big (1-M_2^{(1)}\beta_{21}    -M_3^{(1)} (\beta_{32}\beta_{21}+\beta_{31})       \big )\epsilon_1-(M_2^{(1)}+M_{3}^{(1)}\beta_{32})\epsilon_2-M_3^{(1)}\epsilon_3, \\
	e_{2, {\cal N}} &=\big (\beta_{21}-M_1^{(2)}  - M_3^{(2)} (\beta_{32}\beta_{21}+\beta_{31})  \big )\epsilon_1+(1-M_3^{(2)}\beta_{32})\epsilon_2-M_3^{(2)}\epsilon_3, 
\end{align*} 
where $M^{(l)}_k$ denotes the $k$-th element of  $ {\bM}_{\cal N}^{(l)} = -{\mathbf{\Theta}}_{-ll}/{{\Theta}}_{ll}$ and can be regarded as the partial correlation between nodes $k$ and $l$ given all the other nodes in ${\cal N}$. Since $\Theta_{31}=-\sigma_3^{-2}\beta_{31}\neq0$ and $\Theta_{32}=-\sigma_3^{-2}\beta_{32}\neq 0$, $M_3^{(1)}$ and $M_3^{(2)}$ are nonzero, and thus both $e_{1, {\cal N}}$ and $e_{2, {\cal N}}$ are dependent with $x_3$, leading to ${\cal A}_{0}=\{3,4\}$ and ${{\cal S}_1=\{1,2\}}$. It also follows from Corollary \ref{cor:parents}  that $\mbox{pa}_3=\{1,2\}$ and $\mbox{pa}_4=\emptyset$.

Next, we repeat the above procedure for $x_1$ and $x_2$, and obtain $e_{1, {\cal S}_1}=(1-M_{2}^{(1)}\beta_{21})\epsilon_1- M_2^{(1)}\epsilon_2 \ ~\mbox{and}~\  e_{2, {\cal S}_1}=\epsilon_2$. Clearly,  $e_{2, {\cal S}_1}$ is independent with $x_1$, whereas $e_{1, {\cal S}_1}$ is dependent with $x_2$ due to the fact that $M_{2}^{(1)}\neq 0$.  Therefore, we have ${\cal A}_{1}=\{2\}$ and $\mbox{pa}_2=\{1\}$. Finally, the last remaining node $1$ is assigned to  ${\cal A}_2$, and the topological layers and directed structure of the DAG in \eqref{sec2:toy} are perfectly reconstructed.

\section{DAG learning algorithm}\label{sec:alg}

Given a sample matrix $\bX=\ ( {\bx}_{1}, ...,   {\bx}_{p} )  \in {\cal R}^{n\times p}$ with ${\bx}_l=(x_{1l},...,x_{nl})^T$, we first obtain the  estimated precision matrix  $\widehat{\mathbf{\Theta}}$, and then the residual is $\widehat{\be}_{l,{\cal N}}={\bx}_l + {\bX}_{-l}  \widehat{\mathbf{\Theta}}_{-ll}/\widehat{{\Theta}}_{ll}$, where $ {\bX}_{-l}$ denotes the sample matrix without ${\bx}_l$. 
To invoke Theorem \ref{thm:A0}, we proceed to test the independence between $\widehat{\be}_{l,{\cal N}}$ and all the nodes in ${\cal N}\setminus \{l\}$, and estimate ${\cal A}_0$ as
$$
\widehat{\cal A}_0=\Big \{l:   \widehat{\be}_{l, {\cal N}} ~\mbox{is tested to be independent with} \ x_k \ \mbox{for any} \ k \in {\cal N}\backslash  \{ l \}\Big \}.
$$
Let $\widehat{\cal S}_1={\cal N}\backslash \widehat{\cal A}_0$, then Corollary \ref{cor:parents} implies that for each $l \in \widehat{\cal A}_0$,
$$
\widehat{\mbox{pa}}_l=\Big \{k\in \widehat{\cal S}_1:   \widehat{\Theta}_{lk}\neq 0 \Big \}, 
$$
and $\widehat{\beta}_{lk}=-\widehat{\Theta}_{lk}/ \widehat{\Theta}_{ll}$ for any $k \in \widehat{\mbox{pa}}_l$.

Suppose that the estimated layers $\widehat{\cal A}_{0},...,\widehat{\cal A}_{t-1}$ are obtained, we denote $\widehat{\cal S}_{t}={\cal N}\backslash \{ \cup_{d=0}^{t-1}\widehat{\cal A}_{d}\}$ and estimate the corresponding precision matrix $\widehat{\mathbf{\Theta}}^{\widehat{\cal S}_t}$. The residuals can be computed as $\widehat{\be}_{l, \widehat{\cal S}_t}=\bx_l + \bX_{\widehat{\cal S}_t \backslash \{l\}} [\widehat{\mathbf{\Theta}}^{{\cal S}_t}]_{-l l}/[\widehat{{\Theta}}^{{\cal S}_t}]_{ll}$ for any $l\in \widehat{\cal S}_{t}$, where $\bX_{\widehat{\cal S}_t \backslash \{l\}}$ denotes the sample matrix corresponding to $\widehat{\cal S}_t \backslash \{l\}$. By Corollary \ref{cor:layers},  ${\cal A}_{t}$ can be estimated as
$$
\widehat{\cal A}_{t}=\Big \{l:   \widehat{\be}_{l, \widehat{\cal S}_{t}} ~\mbox{is tested to be independent with }~ x_k ~\mbox{for any}~ k\in \widehat{\cal S}_{t}\backslash \{l\} \Big \}.
$$
Let $\widehat{\cal S}_{t+1}={\cal N}\backslash \{ \cup_{d=0}^{t}\widehat{\cal A}_{d}\}$, then Corollary \ref{cor:parents} implies that for each $l \in \widehat{\cal A}_{t}$, 
$$
\widehat{\mbox{pa}}_l=\Big \{k\in \widehat{\cal S}_{t+1}:   [\widehat{{\Theta}}^{ \widehat{\cal S}_{t}}]_{lk}\neq 0 \Big \},
$$
and $\hat{\beta}_{lk}=-[\widehat{{\Theta}}^{ \widehat{\cal S}_{t}}]_{lk}/[\widehat{{\Theta}}^{ \widehat{\cal S}_{t}}]_{ll}$, for any $k \in \widehat{\mbox{pa}}_l$.  The procedure is repeated until  $|\widehat{\cal S}_t|\leq1$, and we set $\widehat{T}=t+1$ and $\widehat{\cal A}_{t}=\widehat{\cal S}_t$ if  $|\widehat{\cal S}_t|=1$, and $\widehat{T}=t$ otherwise. 

The details of the developed DAG learning method is summarized in Algorithm \ref{alg:2}.

\begin{singlespace}
	\begin{algorithm}[htb]
		\caption{}
		\label{alg:2}
		
		\begin{algorithmic}[1]
			\STATE \textbf{Input: } $\bX \in {\cal R}^{n\times p}$, $\widehat{\cal S}=\{1,...,p\}$, $\widehat{\bB}=\{\hat{\beta}_{ij}\}_{i,j=1}^p=\mathbf{0}_{p\times p}$ and $t=0$.
			\STATE  {\bf Repeat:} until   $|\widehat{\cal S}|\leq1$:
			\begin{itemize}
				\setlength\itemsep{-0.5em}
				\item[a.]   Estimate $\widehat{\mathbf{\Theta}}^{\widehat{\cal S}}$ and compute $\widehat{\be}_{l, \widehat{\cal S}}$ for any $l\in \widehat{\cal S}$;
				\item[b.]	Estimate  $\widehat{\cal A}_{t}=\Big \{l: \widehat{\be}_{l, \widehat{\cal S}} \indep  x_k, ~\mbox{for any}~ k\in \widehat{\cal S}\backslash \{l\} \Big \}
				$;
				\item[c.]  Estimate $\widehat{\mbox{pa}}_{l}=\{k \in \widehat{\cal S}\backslash \widehat{\cal A}_t: [\widehat{{\Theta}}^{\widehat{\cal S}}]_{kl}\neq 0 \}$ and $\widehat{\beta}_{lk}=-[\widehat{{\Theta}}^{\widehat{\cal S}}]_{lk}/[\widehat{{\Theta}}^{\widehat{\cal S}}]_{ll}$ for any $l \in \widehat{\cal A}_{t}$ and $k \in \widehat{\mbox{pa}}_l$;
				\item[d.] Let $\widehat{\cal S}=\widehat{\cal S}\backslash \widehat{\cal A}_{t}$ and $t\leftarrow t+1$.
			\end{itemize}
			\STATE		{ If} $|\widehat{\cal S}|=1$,  set $\widehat{T}=t+1$ and $\widehat{\cal A}_{t}=\widehat{\cal S}$; otherwise set $\widehat{T}=t$.
			\STATE	{\bf Return:} $\{\widehat{\cal A}_t\}_{t=0}^{\widehat{T}-1}$ and  $\widehat{\bB}$.
		\end{algorithmic}
	\end{algorithm}
\end{singlespace}

Note that the performance of the proposed method relies on the accuracy of  the precision matrix estimation in Step 2a and the independence test procedure in Step 2b. Many existing methods in literature can be adopted, such as the  graphical Lasso algorithm \citep{Friedman2008, Ravikumar2011} or the  constrained sparse estimation method \citep{Cai2011} for precision matrix estimation, and the Hilbert-Schmidt independence criterion \citep{Gretton2008}, the distance covariance measure \citep{Szekely2007, Szekely2009} or the ball divergence \citep{PanWL2018} for independence test. For illustration, we adopt the graphical Lasso algorithm and the distance covariance measure in the proposed method, which yields satisfactory performance in all the numerical experiments in Section 6.



\subsection{Computational complexity}\label{sec:alg:complexity}

The computational complexity of  Algorithm \ref{alg:2} is largely determined by the precision matrix estimation and independence tests in Steps 2a and 2b. Particularly, the complexity of Step 2a by using the graphical Lasso algorithm \citep{Friedman2008}  is of order $O\big ((\sum_{k=t}^{T-1} |{\cal A}_k|)^3 \big )$ with $|{\cal A}_t|$ being the cardinality of ${\cal A}_{t}$, and the complexity of Step 2b by using the distance covariance measure \citep{Szekely2007, Szekely2009}  is $O\Big (n^2    (\sum_{k=t}^{T-1}|{\cal A}_k|) (\sum_{k=t}^{T-1}|{\cal A}_k|-1) \Big)$. Therefore, the computational complexity of Algorithm \ref{alg:2} in learning a random  linear non-Gaussian DAG with $T$ layers  is of order  $O\Big( \sum_{t=0}^{T-1}\big ((\sum_{k=t}^{T-1}|{\cal A}_k|)^3  +n^2   (\sum_{k=t}^{T-1} |{\cal A}_k| )(\sum_{k=t}^{T-1}|{\cal A}_k|-1)  \big ) \Big)$. In the worst-case scenario with $T=p$, the computational complexity of Algorithm \ref{alg:2} becomes $O\Big ( p^4+n^2 p^3 \Big )$; and when the DAG is a shallow hub graph with $T=2$, the computational complexity of Algorithm \ref{alg:2} becomes $O\Big ( p^3+n^2 p(p-1) \Big )$.  

It is important to remark that the computational complexity of  the proposed method  is significantly less than most existing methods for learning linear non-Gaussian DAG. For example, the MDirect method \citep{Wang2020} is one of the most recently proposed methods in literature, and its computational complexity in the worst-case scenario is at least of order $O( J^2(n+J)p^{J+1})$, where  $J$ denotes the maximum  in-degree of the DAG. It is of an exponential order in $J$, and thus  MDirect may suffer serious computational challenges when some nodes have a relatively large number of parents.

\section{Statistical guarantees} \label{sec:theory}

In this section,  we establish asymptotic consistency of the proposed method with the graphical Lasso and distance covariance measure in terms of exact DAG recovery. The consistency results are established with explicit dependence on the sample size $n$,  the number of nodes $p$, and the maximum cardinality of the Markov blankets $d=\max_{l \in {\cal N}} |\mbox{mb}_l|$.  Note that the support  of $\mathbf{\Theta}_{-ll}$ is a subset of the Markov blanket of node $l$ \citep{Park2021}.

For simplicity, denote $\sigma_{max}^2=\max_{l \in {\cal N}} \sigma_l^2$, $\sigma_{min}^2=\min_{l \in {\cal N}} \sigma_l^2$, $\beta_{max}=\max_{l \in {\cal N}, k\in \mbox{pa}_l} |\beta_{lk}|$ and $\beta_{min}=\min_{l \in {\cal N}, k\in \mbox{pa}_l} |\beta_{lk}|$. Further, denote $f(n) =\Omega( g(n))$ if there exists a positive constant $a$ such that $f(n) \geq ag(n)$ for all sufficiently large $n$. The following technical assumptions are made to establish the exact DAG recovery.

\begin{Assumption} \label{ass::incoh}
	There exists some constant $\psi \in (0,1]$ such that 
	\begin{align*}
		\max_{t\in \{0,...,T-1\}} \max_{r \in {\cal C}_t^c} \| \Gamma_{r {\cal C}_t} (\Gamma_{{\cal C}_t {\cal C}_t})^{-1} \|_1 \leq 1-\psi,
	\end{align*}
	where $\Gamma=\mathbf{\Sigma} \otimes \mathbf{\Sigma}$ with $\otimes$ denoting the Kronecker product,  $\Gamma_{(l,k),(j,m)}:=\Gamma_{lp+k,jp+m}=\Sigma_{lk} \Sigma_{jm}$, ${\cal C}_t=\{ (l, k) \in {\cal S}_t \times {\cal S}_t : [{\Theta}^{{\cal S}_t}]_{lk} \neq 0\}$ and ${\cal C}_t^c=({\cal S}_t \times {\cal S}_t)\backslash {\cal C}_t$.
\end{Assumption}
\begin{Assumption} \label{ass::noise}
	For any $j\in {\cal N}$, $\epsilon_j/\sigma_j$ follows a sub-Gaussian distribution with parameter $\gamma$.
\end{Assumption}

Assumption \ref{ass::incoh} limits the correlation between the zero and non-zero elements in $\Gamma$, which is analogous to the irrepresentable condition in \cite{Zhao2006} or the incoherence condition in \cite{Ravikumar2011}.   Assumption \ref{ass::noise} characterizes the noise distribution and implies that ${x_j}/{\sqrt{\Sigma_{jj}}}$ also follows a sub-Gaussian distribution with parameter $\gamma$.


\begin{lem}\label{lem::coef}
	Suppose Assumptions \ref{ass::incoh} and \ref{ass::noise} hold, and $n=\Omega(d^2 \log p)$. For any $l\in {\cal N}$, there exist some positive constants $a_1, a_2$ and $\tau>4$ such that with probability at least $1-a_2 p^{2-\tau}$, there holds 
	\begin{align}
		\Big \|\frac{\widehat{\mathbf{\Theta}}_{\cdot l}}{\widehat{\Theta}_{ll}} - \frac{\mathbf{\Theta}_{\cdot l}}{\Theta_{ll}} \Big \|_{2} \leq a_1 \gamma^2 \tau^{1/2} \sqrt{\frac{d\log p}{L_2^2n}},
	\end{align}
	provided that the regularization parameter in estimating $\mathbf{\Theta}$ is $\lambda_{n,0}\propto \sqrt{\log p/n}$, where $L_2=\sigma_{max}^{-2}\min \big \{1 , (\frac{\sigma_{min}}{\sigma_{max}})^2 L_1^{-1} \big \}$ with $L_1=\beta_{max} (1+d \beta_{max})$.
\end{lem}

Lemma \ref{lem::coef} paves a bridge between the estimated precision matrix and the estimated residuals, and plays a crucial role in establishing the consistency for the exact DAG recovery.

\begin{Assumption} \label{ass::depen}
	There exists a positive constant $\lambda_{max}$ such that
	$
	\Lambda_{max} \big ( \frac{1}{n} {\bX}^T {\bX} \big ) \leq \lambda_{max},
	$
	where $\Lambda_{max}(\cdot) $ denotes the maximum eigenvalue of  a matrix.
\end{Assumption}

\begin{Assumption}\label{ass::dcov} 
	For any $t=0,..., T-1$, we have
	\begin{equation*}
		\max_{k \in {\cal S}_{t}\backslash \{l\}} \mbox{dcov}^2 (e_{l,{\cal S}_{t}}, x_k)\left \{
		\begin{aligned}
			&=0, &\text{if}& \ l\in {\cal A}_t; \\
			&\geq \rho_{n,t}^2,  &\text{if}& \ l\in {\cal S}_t \backslash {\cal A}_t,
		\end{aligned}
		\right.
	\end{equation*}	
	where
	$\rho_{n,t}^2=\Omega(\max\{ d^3 n^{-1/2} \log^{1/2}(\max \{|{\cal S}_t|,n\}),n^{-\eta}\})$ with $0<\eta<\frac{1}{2}$, and $|{\cal S}_{t}|$ denotes the cardinality of ${\cal S}_t$.
\end{Assumption}

Assumption \ref{ass::depen} is a standard regularity condition on the sample covariance matrix of $\bX$ in ${\cal N}$, which also regulates the sample covariance matrix of $\bX_{{\cal S}_t}$ since $\Lambda_{max}\big (\frac{1}{n} {\bX}_{{\cal S}_t}^T {\bX}_{{\cal S}_t}\big ) \leq 	\Lambda_{max} \big ( \frac{1}{n} {\bX}^T {\bX} \big )$ for any $t=0,...,T-1$. Assumption \ref{ass::dcov} assures that the distance covariance is sufficient in discriminating nodes in ${\cal A}_t$ or ${\cal S}_t \backslash {\cal A}_t$. Similar assumptions have also been employed in \cite{Kalisch2007} and \cite{Ha2016}.

\begin{theorem}\label{thm::consis::A0}
	\noindent {\bf(Consistency of $\widehat{{\cal A}}_0$)}  Suppose that  all the assumptions in Lemma \ref{lem::coef} and Assumptions \ref{ass::depen} and \ref{ass::dcov} hold, and $n = \Omega ( d^6 \log p )$. Then there exist some positive constants $a_3,a_4$ and $a_5$ such that 
	\begin{align*}
		P(\widehat{\cal A}_0 = {\cal A}_0) \geq 1-  a_{3} p^{4-\tau} - a_{4} p^2 \exp\{-a_{5} n^{(1-2\eta)/3}\},
	\end{align*}
	provided that the significant level of the independence test is set as $\alpha_n$, and $\alpha_n \rightarrow 0$ as $n$ diverges.
\end{theorem}

Theorem \ref{thm::consis::A0} shows that the lowest layer  ${\cal A}_0$ in ${\cal G}$ can be exactly recovered by the proposed method with high probability.  After reconstructing ${\cal A}_0$,  the selection consistency of the parent set for nodes in ${\cal A}_0$ can also be established, following from  the fact that $\min_{l \in {\cal A}_0, k \in \mbox{pa}_l}|\Theta_{lk}| = \min_{l \in {\cal A}_0, k \in \mbox{pa}_l}\sigma_l^{-2}|\beta_{lk}| \geq \sigma_{max}^{-2} \beta_{min}$ and a similar treatment in Theorem 2 of \cite{Ravikumar2011}. 

\begin{cor}\label{cor::consis::a0::pa}
	Suppose that all the assumptions in Theorem \ref{thm::consis::A0} are satisfied, and $n = \Omega ( (d^6 + \beta_{min}^{-2})  \log p)$. Then there exists some positive constant $a_{6}$ such that 
	$$
	P \Big ( \big \{\widehat{\mbox{pa}}_l = \mbox{pa}_l : l \in \widehat{\cal A}_0 \big \} \Big | \widehat{\cal A}_0= {\cal A}_0\Big ) \geq 1-a_{6}p^{2-\tau}.
	$$
\end{cor}

Further, let $\widehat{{\cal S}}_1 = {\cal N}\backslash \widehat{\cal A}_0$, and we apply the similar treatment on $\widehat{{\cal S}}_1$ to establish consistency in estimating $\widehat{\cal A}_1$, as well as other upper layers. In the spirit of mathematical induction, we arrive at the following theorem on the asymptotic estimation consistency of $\widehat{\cal G}$.

\begin{theorem}\label{thm:2}
	{\noindent \bf(Consistency of $\widehat{\cal G}$)}  
	Suppose that all the assumptions in  Corollary \ref{cor::consis::a0::pa} are satisfied, and $n=\Omega \Big(T^{1/(\tau-4)}(d^6 + \beta_{min}^{-2})  (\log (\max\{p,n\}))^{3/(1-2\eta)}\Big)$. 
	Then there holds
	$$
	P(\widehat{\cal G}={\cal G}) \longrightarrow 1, \ \mbox{as $n \rightarrow \infty$,}
	$$
	provided that the regularization parameter in estimating $\mathbf{\Theta}^{{\cal S}_t}$ is $\lambda_{n,t}\propto \sqrt{\log (\max\{|{\cal S}_t|,n\})/n}$.
\end{theorem}

Theorem \ref{thm:2} ensures that the linear non-Gaussian DAG ${\cal G}$ can be consistently recovered by the proposed method even under the high dimensional setting. Specifically, with all other terms fixed, the  consistency of exact DAG recovery holds true when $\log(p)=o(n^{(1-2\eta)/3})$ with $0<\eta<1/2$. This is in sharp contrast with the result in \cite{Wang2020} with $\log(p)=o(n^{1/(2K)})$, where $K\geq 3$ denotes the order of moment-based statistic in \cite{Wang2020}. 

Theorem \ref{thm:2} can be further extended by relaxing the sub-Gaussian noise assumption, such as a noise distribution with ($4m$)-th bounded moment; that is, $\max_j E((\epsilon_j/\sigma_j)^{4m}) \leq K_m$ for a positive integer $m$ and a postive constant $K_m$.

\begin{cor} \label{cor::graph::4m}
	Suppose that all the assumptions in  Theorem \ref{thm:2} are satisfied, except that Assumption 2 is relaxed to a noise distribution with ($4m$)-th bounded moment, Assumption 4 holds with $\rho_{n,t}^2=\Omega \big (\max\{d^3 n^{-\frac{1}{2}} |{\cal S}_t|^{\frac{2}{m}} (\max\{|{\cal S}_t|,n\} )^{\frac{\tau-4}{2m}} , n^{-\eta} \}  \big )$ for some constants $4<\tau<m+4$ and $0<\eta<\frac{1}{2}$,   and $n=\Omega \Big( T^{\frac{1}{\min\{\tau-4, 2m\phi-1\}}} (d^6+\beta_{min}^{-2})p^{\max\{\frac{4}{m},\frac{2}{2m\phi-1}\}} (\max\{p,n\})^{\frac{\tau-4}{m}}  \Big)$ for some constant $\frac{1}{2m} < \phi < \frac{1}{2}-\eta$. Then,   there holds 
	$$
	P(\widehat{\cal G}={\cal G}) \longrightarrow 1, \ \mbox{as $n \rightarrow \infty$,}
	$$
	provided that $\lambda_{n,t}\propto |{\cal S}_t|^{\frac{2}{m}} n^{-\frac{m+4-\tau}{2m}}$.
\end{cor}

Corollary \ref{cor::graph::4m} establishes the asymptotic DAG recovery of the proposed method with ($4m$)-th bounded moment noise distributions, which requires a relatively larger sample size compared with that in Theorem \ref{thm:2}. More importantly, both Theorem \ref{thm:2} and Corollary \ref{cor::graph::4m} are established without  assuming the parental faithfulness assumption in \cite{Wang2020}, indicating a more general applicability of the proposed method.

\section{Numerical experiments}\label{sec:num}

In this section, we examine the numerical  performance of the proposed method, denoted as TL, and compare it  against some popular DAG learning  methods, including  the direct high-dimension learning algorithm (MDirect, \cite{Wang2020}),  the pairwise learning  algorithm (Pairwise, \cite{Shimizu2011,Hyvarinen2013}),   the ICA-based learning  algorithm (ICA, \cite{Shimizu2006}),  the high dimensional constraint-based  PC algorithm  (PC, \cite{Kalisch2007}) and a hybrid version of  max-min hill climbing algorithm (MMHC, \cite{Tsamardinos2006}). Particularly, TL adopts the graphical Lasso algorithm with a fixed regularization parameter as suggested in Section 5, and the independence test based distance covariance measure with a significance level $\alpha=0.01$. MDirect is implemented in the  \texttt{R} package \texttt{highDLingam} \citep{Wang2020}, and both the methods ICA and MMHC are implemented in the \texttt{R} package \texttt{CompareCausalNetworks}. Furthermore, we implement Pairwise by using the \texttt{R} package \texttt{causalXtreme} \citep{Gnecco2019}, and further extend it with the Lasso algorithm for DAG in high dimensional cases, and we implement PC by using the \texttt{R} package \texttt{pcalg} \citep{Kalisch2012}, which outputs a partial DAG, and then we apply the treatment in \cite{Yuan2019} to convert it to a DAG by using the \texttt{pdag2dag} routine in the R package \texttt{pcalg}.  Note that the significant level of independent tests in both  TL and PC is set to $\alpha=0.01$,  and the  least square estimation is also applied  to estimate the connection strength of directed structures for MDirect, MMHC and PC.

The numerical performance of all the methods is evaluated in terms of estimation accuracy of   directed edges and coefficients. For  the accuracy of estimated directed edges,  we employ the true positive rate (TPR) and false discovery rate (FDR) as the evaluation metric. To evaluate the closeness of the estimated and true DAG,   we  report the normalized structural Hamming distance \citep{Tsamardinos2006}, which measures the smallest number of edge insertions, deletions, and flips to convert the estimated DAG into the truth. For overall accuracy of  the estimated DAG structure, we use the  Matthews correlation coefficient (MCC) as an overall evaluation metric, which is also considered in \cite{Yuan2019}. For  evaluation of the coefficient estimation, we report the relative error between the estimated  adjacency matrix $\widehat{\bB}$ and the true  adjacency matrix ${\bB}$  in  Frobenius norm that $\text{rel-Fnorm}:= {\| \widehat{\bB}-{\bB}\|_F}/{\|  {\bB} \|_F}$. Note that a good estimation is implied with small values of FDR, HM and rel-Fnorm, but large values of TPR and MCC.

\subsection{Simulated examples}\label{sec:sim}

In this section, the numerical performance of  all the methods are evaluated in two simulated examples, where Examples 1 considers a hub graph,  and Example 2 considers a scale-free graph generated by the   Barab\'asi-Albert (BA) model. 

\begin{figure}[!h]
	\centering
	\begin{subfigure}[b]{0.4\textwidth}
		\includegraphics[width=\textwidth,height=3cm]{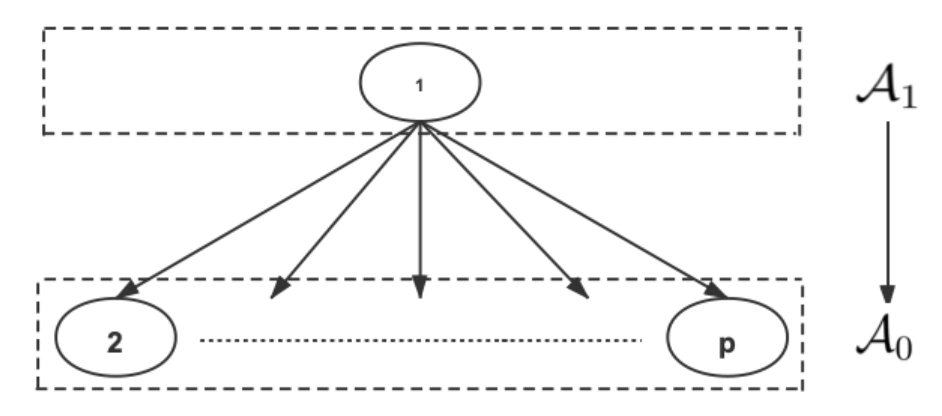}
		\caption{}
		\label{fig:ex11}
	\end{subfigure}
	\hspace{5mm}
	\begin{subfigure}[b]{0.4\textwidth}
		\includegraphics[width=\textwidth]{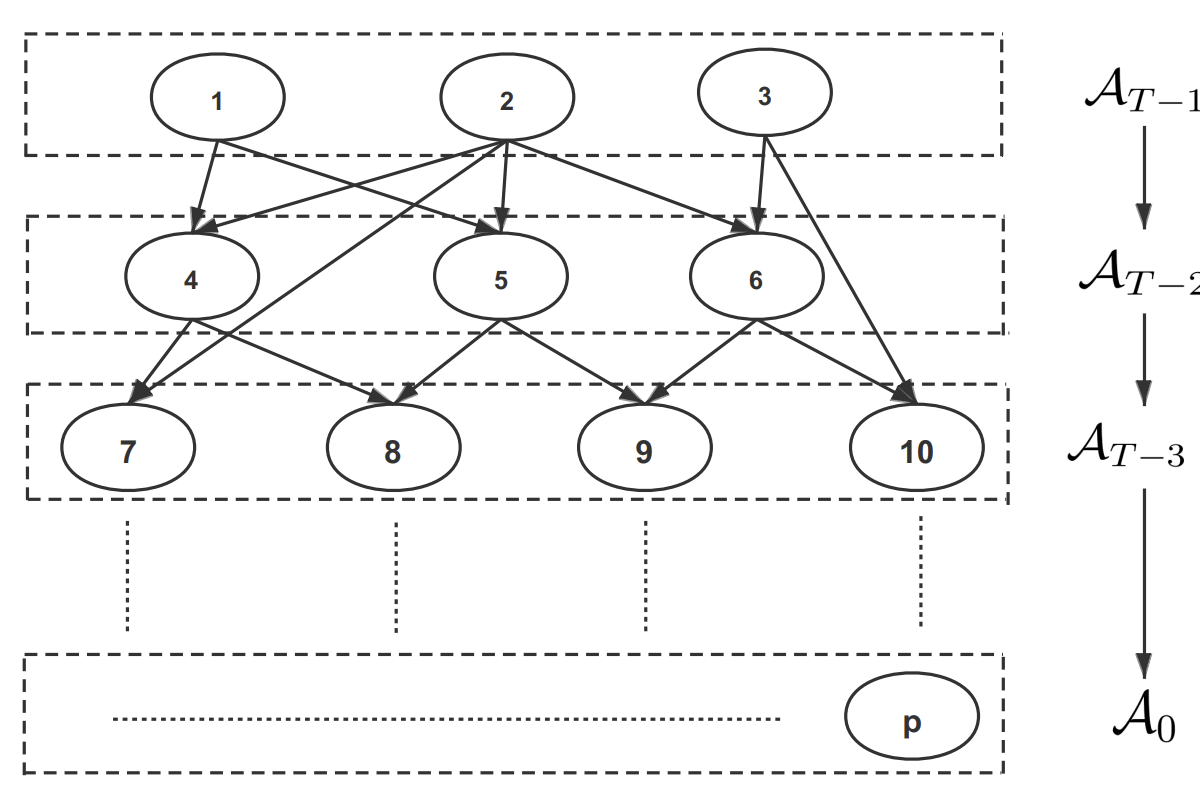}
		\caption{}
		\label{fig:ex13}
	\end{subfigure}
	\caption{The topological layer of the DAG structures in Examples 1 and 2.}
\end{figure}

{\bf Example 1.} We consider a hub graph with $T=2$, ${\cal A}_0=\{2,...,p\}$ and ${\cal A}_1=\{1\}$, whose DAG structure is illustrated in Figure \ref{fig:ex11}.  Moreover, we consider various noise distributions, including uniform distribution on  $[-3, 3]$, student $t$ distribution with $9$ degree of freedom, and double exponential distribution with location parameter 0 and scale parameter $\sqrt{1.5}$, and  the coefficient of each directed edge is uniformly generated from $[-1.5,-0.5]\cup[0.5,1.5]$.

{\bf Example 2.} We consider a similar data generating scheme as in \cite{Wang2020}. Specifically, we start with a graph with only one node, and  at each step, a node with $2$ directed edges are added to the graph. The probability of generating a directed edge from each previous node to the newly added node is proportional to the number of neighbors of the previous node. Its DAG structure is illustrated in Figure \ref{fig:ex13}. Moreover,  we generate the noise terms from $\sigma \times \text{Uniform}[-3,3]$ with $\sigma \sim \text{Uniform}[0.2,1]$, and the coefficient of each directed edge is uniformly generated from $[-1.5,-0.5]\cup [0.5,1.5]$.  

For each example, we repeat the data generating scheme 50 times and  the averaged performance of  all the  methods under the cases with $(n,p)= (200,100), (200, 200),  (400,200)$ and $(400,1000)$ are summarized in Tables  \ref{tab:1} and \ref{tab:3}. Note that   ICA  is only designed for the   low dimensional case with $p<n$,  and some methods do not produce any results for cases with large $p$ in Examples 1 and 2 after more than 48 hours.  

\begin{table}[!h]
	\scriptsize
	\centering
	\begin{tabular}{ccccccc}
		\hline
		$ (n, p)$& Method &  TPR  & FDR   & MCC & HM & rel-Fnorm \\\hline
		$(200,100)$ 
		& TL &0.8657 (0.0060) & 0.0091 (0.0013)& 0.9252  (0.0032)& 0.0014  (0.0001)& 0.2533 (0.0077)\\
		& MDirect &	0.0412 (0.0013) & 0.9907 (0.0002) & -0.0011 (0.0005)& 0.0527 (0.0005)& 1.1335 (0.0008)\\
		& Pairwise &  	0.9564 (0.0017)&  0.6700   (0.0023)& 0.5551 (0.0023)&0.0201   (0.0002) &0.4269 (0.0014)\\
		& ICA& 	0.5020  (0.0030)&0.6506   (0.0019 )&0.4115 (0.0021) &0.0144   (0.0001)& 0.9454 (0.0011)\\
		& MMHC    &  	0.1568  (0.0006)& 0.8289   (0.0009)& 0.1557 (0.0007)& 0.0161  (0.0000)&  1.0006  (0.0013)\\
		& PC  & 	0.1107  ( 0.0008) & 0.6608  (0.0035)&0.1890  (0.0015)&0.0111  (0.0000) & 0.9395  (0.0010)\\
		\hline
		$(200,200)$ 
		& TL & 0.9267  (0.0023) & 0.0099 (0.0012)& 0.9576 ( 0.0013)& 0.0004 (0.0000)&0.1812 (0.0039)\\
		& MDirect &		 0.0179  (0.0005)  &0.9963 (0.0001)   &-0.0027 (0.0002)   &0.0286 (0.0002) & 1.1541 (0.0015) \\
		& Pairwise  &  0.8967  (0.0017) & 0.7982  (0.0016) &  0.4201 (0.0019)& 0.0185  (0.0002)&  0.5109  (0.0013)\\
		& ICA& ** &   ** &   ** &  ** & ** \\
		& MMHC    & 	0.0817 (0.0004)&  0.9234  (0.0004) &  0.0743  (0.0004)&0.0095 (0.0000) &   1.0823 (0.0008) \\
		& PC  &	0.0565 (0.0004) & 0.8646  (0.0013) & 0.0844 (0.0007) &0.0065 (0.0000)&  0.9974  (0.0005)\\
		\hline 
		$(400,200)$ 
		& TL & 	0.8543 (0.0037)  & 0.0002 (0.0002) &  0.9237 (0.0020) & 0.0007 (0.0000) &0.2566 (0.0037)\\
		& MDirect &   	0.0175 (0.0004) &0.9963 (0.0001) &  -0.0026  (0.0002) &0.0276  (0.0002)&1.1312 (0.0003)\\
		& Pairwise  &	0.9819 (0.0005)  &0.6890  (0.0014) & 0.5490 (0.0013)&0.0111 (0.0001) & 0.3541 (0.0005) \\
		& ICA&   0.5169  (0.0017) &0.7392  (0.0009)& 0.3627 (0.0012) &0.0098 (0.0000)&   0.9484  (0.0005)\\
		& MMHC    & ** &   ** &   ** &  ** & ** \\
		& PC  &  ** &   ** &   ** &  ** & ** \\
		\hline
		$(400,1000)$ 
		& TL &  0.9633 (0.0008) & 0.0003 (0.0001)& 0.9813 (0.0004)& 0.0000 (0.0000)& 0.1269 (0.0015)\\
		& MDirect &   ** &   ** &   ** &  ** & ** \\
		& Pairwise  &  ** &   ** &   ** &  ** & ** \\
		& ICA& ** &   ** &   ** &  ** & ** \\
		& MMHC    & ** &   ** &   ** &  ** & ** \\
		& PC  &   ** &   ** &   ** &  ** & ** \\
		\hline
	\end{tabular}
	\caption{The averaged  measures of all the methods in Example 1 together with their standard errors in parentheses. Here ** denotes the fact that the corresponding methods are either not applicable or take too long to produce any results.}
		\label{tab:1}
\end{table}

\begin{table}[!h]
	\scriptsize
	\centering
	\begin{tabular}{ccccccc}
		\hline 
		$ (n, p)$& Method &  TPR  & FDR   & MCC & HM & rel-Fnorm \\\hline  
		$(200,100)$ & TL &  0.6419 (0.0083) & 0.1883  (0.0096)&  0.7167 (0.0086) & 0.0101(0.0003) &0.7717 (0.0235)\\
		& MDirect &   0.1267  (0.0015) &  0.9147 (0.0011)  &0.0818  (0.0013)  &  0.0446 (0.0002) & 1.1006 (0.0030)\\
		& Pairwise  & 0.3192  (0.0043) & 0.9395  (0.0011)  & 0.0994  (0.0024) & 0.1140  (0.0008) &   0.9446 ( 0.0050) \\
		& ICA&   0.7202  (0.0028) & 0.7019  (0.0013) & 0.4474 (0.0013)  & 0.0395  (0.0002)  &    0.8246  (0.0017)\\
		& MMHC    &  0.3124 ( 0.0022)  & 0.3225  (0.0047) & 0.4530  (0.0031)  & 0.0167 (0.0001) &    0.8738  (0.0039)\\	& PC  &  0.2064 (0.0024) &  0.5599  (0.0051) & 0.2921  (0.0035)  & 0.0210  (0.0001) &  0.9381 (0.0023)\\
		\hline 
		$(200,200)$	& TL  & 0.6694 (0.0082)& 0.2611 (0.0080)&  0.7002 (0.0077)& 0.0057 (0.0001)& 0.8217 (0.0348)\\
		& MDirect &  0.0923 (0.0009) & 0.9397  (0.0006)  & 0.0631  (0.0007)  &  0.0234  ( 0.0001) & 1.1261 (0.0022)\\
		& Pairwise  &0.3195 (0.0031) &0.9525  (0.0006) &  0.1012  (0.0014) &0.0713  (0.0004) &    0.9267 (0.0016)\\
		& ICA&   ** &   ** &   ** &  ** & ** \\
		& MMHC   & 0.3042  (0.0019) & 0.3121  (0.0034) & 0.4539 (0.0024) &0.0083  (0.0000)  &   0.8771 (0.0026)\\
		& PC  &  0.1897  (0.0022) &  0.5848  (0.0044) &     0.2759  (0.0031)  & 0.0107  (0.0000) &   0.9521 (0.0023)\\
		\hline 
		$(400,200)$ & TL & 0.6751 (0.0073) & 0.1605   (0.0064) & 0.7504 (0.0066) &  0.0045 (0.0001)&0.7075 (0.0183)\\
		& MDirect & 0.0976  (0.0007) &  0.9381 (0.0005) & 0.0660  (0.0006)  &  0.0239 (0.0001) & 1.1163 (0.0017) \\
		& Pairwise  &  0.3155 (0.0029) & 0.9617    (0.0004) & 0.0852  (0.0011) &  0.0865 (0.0003) & 0.9351 (0.0013) \\
		& ICA& 0.8016 (0.0013) &  0.7804  (0.0005) &  0.4098 (0.0005) &  0.0305 (0.0001) & 0.8101 (0.0009) \\
		& MMHC    &  0.3344 (0.0015) & 0.3156 (0.0026)  & 0.4749 (0.0019) & 0.0082 (0.0000) & 0.8588 (0.0018) \\
		& PC  &  0.1993 (0.0019) & 0.6225  (0.0036)  & 0.2691 (0.0026)  &  0.0113 (0.0000) & 0.9501 (0.0021)\\
		\hline
		$(400,1000)$  & TL & 0.6041 (0.0097) & 0.1113  ( 0.0037) &  0.7313 (0.0066) & 0.0009  (0.0000) &0.6752 (0.0091)\\
		& MDirect & 0.0522  (0.0003) &  0.9654 (0.0002) &  0.0401  (0.0002)  &  0.0048 (0.0000) & 1.1422 (0.0010) \\
		& {Pairwise } &  ** &   **  &  **  &   **  & **  \\
		& ICA&  ** &   **  &  **  &   **  & **  \\
		& MMHC    & 0.3041  (0.0009) & 0.3119  (0.0015)  & 0.4567 (0.0012) &  0.0017 (0.0000) & 0.8807 (0.0014) \\
		& PC  &  0.1612 (0.0006) & 0.6818  (0.0012)  & 0.2254 (0.0009)  & 0.0024 (0.0000) &  0.9754 (0.0009)\\
		\hline
	\end{tabular}
	\caption{The averaged  measures of all the  methods in Example 2 together with
	their standard errors in parentheses. Here ** denotes the fact that the corresponding methods are either not applicable or take too long to produce any results.}
\label{tab:3}
\end{table}

It is evident from Tables \ref{tab:1} and \ref{tab:3} that TL outperforms all the other competitors in almost all  the cases, except that it yields the second best TPR in the cases with $(n, p)=(200, 100)$ and $(400, 200)$. In these two cases, Pairwise or ICA attain higher TPR, largely due to the fact that they tend to produce very dense graphs with many false edges and thus have much higher FDR. Note that the performance of MDirect appears less satisfactory, possible due to its sensitivity to the data generating scheme.   It is also interesting to point out that the performance of TL may be further improved with a finer tuning scheme, at the cost of increasing computational cost.

\subsection{Spread of COVID-19}

\begin{figure}[h]
	\centering
	\begin{minipage}[b]{0.45\textwidth}
		\centering
		\includegraphics[width=1\textwidth]{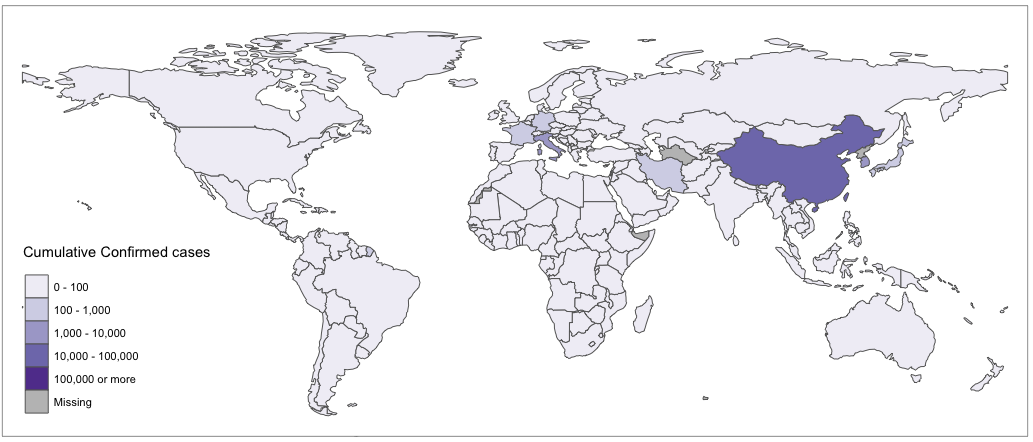}
		\subcaption{March 1, 2020}
		\label{mapFig1}
	\end{minipage}
	\begin{minipage}[b]{0.45\textwidth}
		\centering
		\includegraphics[width=1\textwidth]{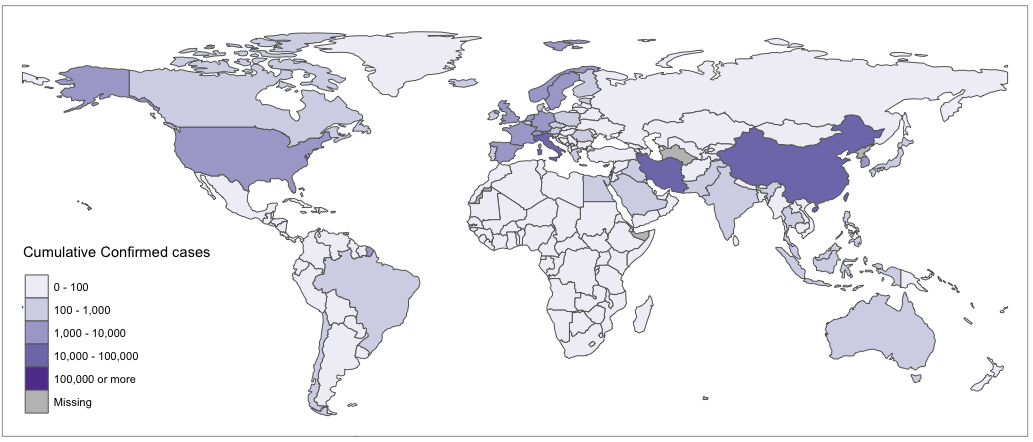}
		\subcaption{March 15, 2020}
		\label{mapFig2}
	\end{minipage}
	\begin{minipage}[b]{0.45\textwidth}
		\centering
		\includegraphics[width=1\textwidth]{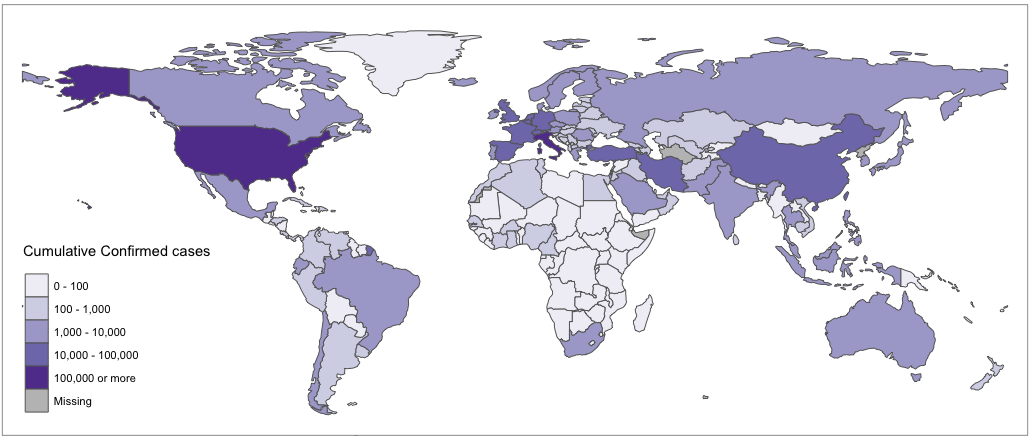}
		\subcaption{March 30, 2020}
		\label{mapFig3}
	\end{minipage}
	\begin{minipage}[b]{0.45\textwidth}
		\centering
		\includegraphics[width=1\textwidth]{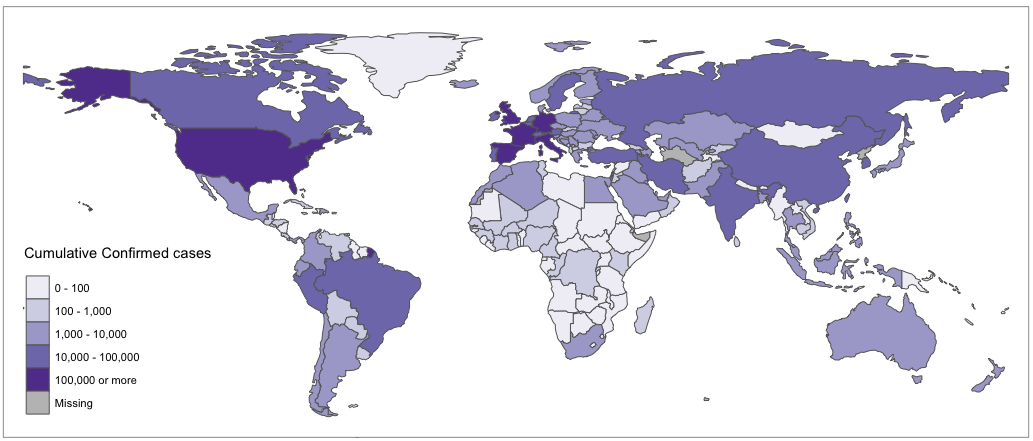}
		\subcaption{April 15, 2020}
		\label{mapFig4}
	\end{minipage}
	\caption{Heat map of the cumulative confirmed cases for countries around the world from March 1st, 2020 to April 15th, 2020.}\label{heatmap}
\end{figure}

We now apply TL to analyze the spread of COVID-19 based on the daily global confirmed cases collected by the Center for Systems Science and Engineering (CSSE) at Johns Hopkins University, which is publicly available at \url{https://github.com/CSSEGISandData/COVID-19}. Figure \ref{heatmap} displays the heat maps for the global cumulative confirmed cases for countries around the world from March 1st, 2020 to April 15th, 2020. It is clear that most confirmed cases of COVID-19 are first reported in China, Europe and  Iran, then it quickly spreads to Middle East and North America, and finally most of the countries are affected by COVID-19, especially USA and west European countries. Interestingly, compared with other continents, countries in Africa appear to be much less affected.

It is interesting to note that DAG is an efficient tool to describe the spread of COVID-19, where a directed edge indicates the virus is spread from one country to the other. Although virus-spread may not be necessarily acyclic, DAG provides insightful information on the future infection tendency of virus-spread among the countries. We pre-process the dataset and exclude those countries or regions with no confirmed cases for more than 10 days during March 1st, 2020 to April 15th, 2020. This leads to the daily confirmed cases in  $p=99$ countries or regions for a total of 61 days. Further, we  convert the actual number of daily confirmed cases to the percentage over all countries, and use a 3-day moving average of the percentages as the observation for each day. We then apply TL to estimate the DAG for the spread of COVID-19, with 99 nodes and 736 directed edges.

\begin{figure}[!h]
	\centering
	\includegraphics[width=0.8\textwidth]{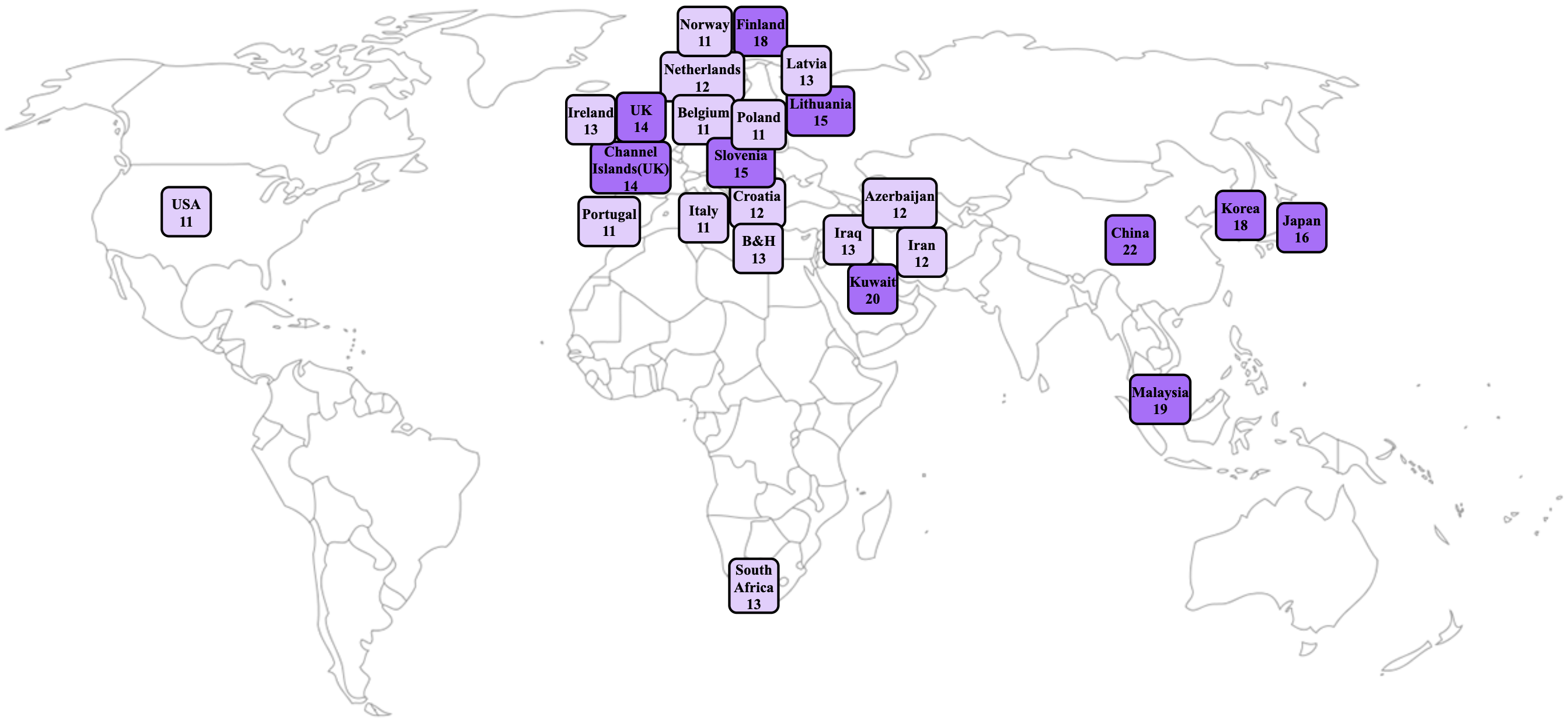}
	\caption{Top  25 hub nodes with the number of their child nodes in the estimated DAG for the spread of COVID-19. }	
	\label{hubnodes}
\end{figure}

\begin{figure}[!h]
	\centering
	\includegraphics[width=0.8\textwidth]{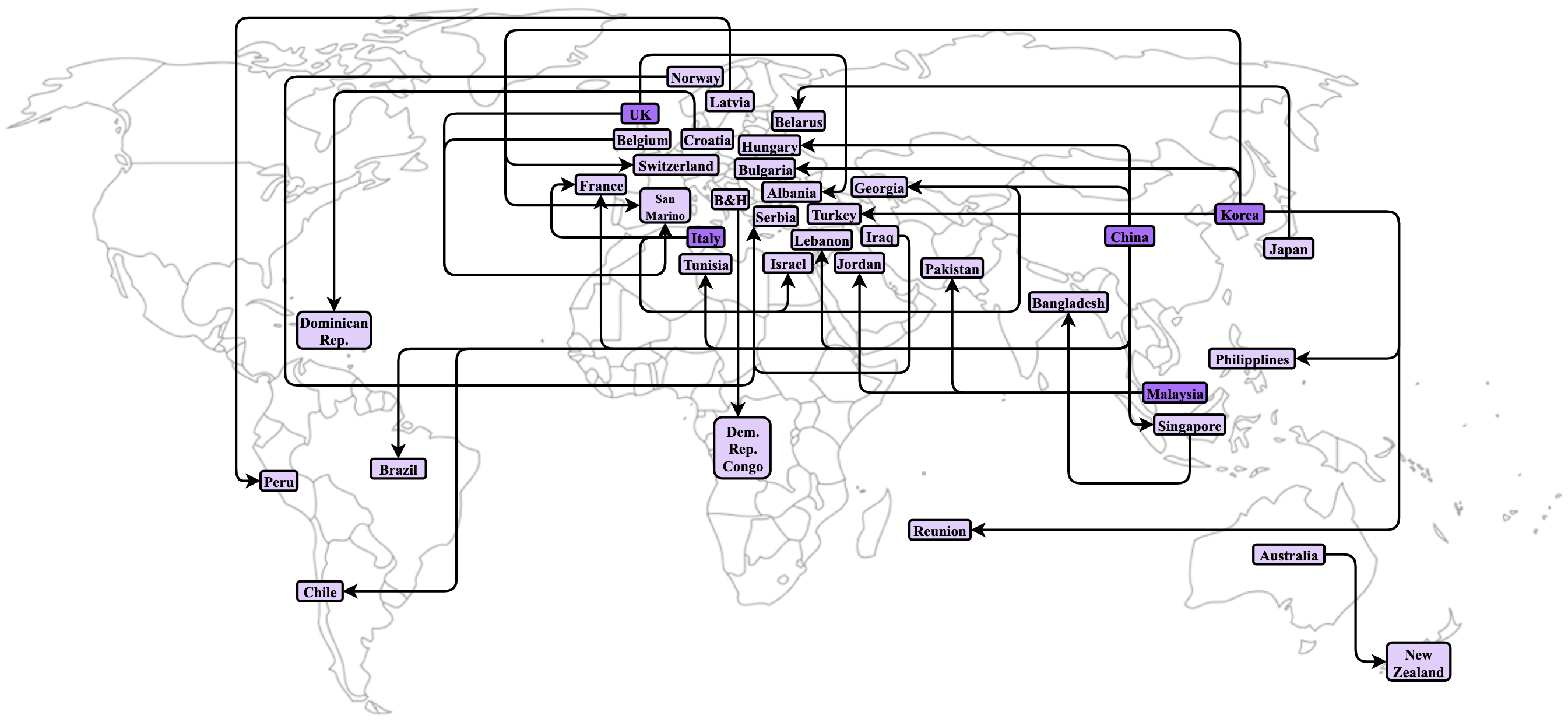}
	\caption{Top 30 directed edges in the estimated DAG for the spread of COVID-19.}
	\label{mapDAGtop30}
\end{figure}

Figure \ref{hubnodes} shows the top 25 hub nodes with the number of their child nodes in the estimated DAG for the spread of COVID-19, which consists of mostly Eastern Asian and Western Asian countries, and most European countries. This concurs with the heat map in Figure \ref{mapFig2} that these countries reported many confirmed cases in middle March, and thus are more likely to spread the virus. Figure \ref{mapDAGtop30} presents 30 directed edges with largest estimated weights, showing that China, Korea, Italy and United Kingdom are the major countries that spread the virus to others. This trend of infection appear sensible, since these countries have more confirmed cases in early March as shown in Figure \ref{heatmap} and they are also closely connected to other countries due to their active economy or tourism attractions. Moreover, many directed edges are present among European countries, largely due to the fact that population movement and interaction in these countries are much more frequently than others. It is also interesting to note that Figure \ref{mapDAGtop30} shows no directed edges point from the United States of America and Canada to other countries, yet they are indeed hub nodes with 11 and 6 child nodes, respectively. Given the fact that there is a surge in the number of confirmed cases in these two countries as observed in Figure \ref{heatmap}, one can expect the virus would spread from these two countries to their child nodes after April, 2020.



\section{Discussion}\label{sec:sum}

This paper proposes an efficient method to learn linear non-Gaussian DAG in high dimensional cases with statistical guarantees. The proposed method leverages a novel concept of topological layers to facilitate DAG learning, which ensures that the parents of a node must belong to its upper layers, and thus naturally guarantees acyclicity. To learn the DAG, its layers can be reconstructed via precision matrix estimation and independence tests in a bottom-up fashion, and its parent-child relations can be directly obtained from the estimated precision matrix.	More importantly, the proposed method can consistently recover the underlying DAG under more mild conditions than existing methods in literature. Its advantages over some popular competitors are also supported by numerical experiments on a variety of simulated and real-life examples.

\acks{	XH's research is supported in part by  NSFC-11901375 and Shanghai Pujiang Program 2019PJC051, and JW's research is supported in part by GRF-11303918, GRF-11300919, and GRF-11304520.}


\appendix

\bibliography{bibtex.bib} 
\bibliographystyle{plainnat}

\end{document}